\documentclass[conference,a4paper]{IEEEtran}
\usepackage{float}

\usepackage{cite}
\usepackage{graphicx}
\ifCLASSINFOpdf
\else
\fi
\usepackage{amsmath}
\usepackage{amssymb}
\usepackage{bm}

\usepackage{pifont}% http://ctan.org/pkg/pifont
\newcommand{\cmark}{\ding{51}}%
\newcommand{\xmark}{\ding{55}}%
\newtheorem{exe}{Theorem}

\newtheorem{definition}[exe]{Definition}
\def\vec#1{{\boldsymbol{#1}}}
\def\mat#1{{\boldsymbol{#1}}}

\def\vec#1{{\boldsymbol{#1}}}
\def\mat#1{{\boldsymbol{#1}}}

\begin{document}
\title{From Review to Rating: Exploring Dependency Measures for Text Classification}

\author{\IEEEauthorblockN{Samuel Cunningham-Nelson}
\IEEEauthorblockA{School of Electrical Engineering \\ and Computer Science\\
Queensland University of Technology\\
Brisbane, Queensland\\
samuel.cunninghamnelson@qut.edu.au}
\and
\IEEEauthorblockN{Mahsa Baktashmotlagh}
\IEEEauthorblockA{School of Electrical Engineering \\ and Computer Science\\
Queensland University of Technology\\
Brisbane, Queensland\\
m.baktashmotlagh@qut.edu.au}
\and
\IEEEauthorblockN{Wageeh Boles}
\IEEEauthorblockA{School of Electrical Engineering \\ and Computer Science\\
Queensland University of Technology\\
Brisbane, Queensland\\
w.boles@qut.edu.au}}


%Telephone: (800) 555--1212, Fax: (888) 555--1212}


\maketitle

\begin{abstract}
Various text analysis techniques exist, which attempt to uncover unstructured information from text. In this work, we explore using statistical dependence measures for textual classification, representing text as word vectors. Student satisfaction scores on a 3-point scale and their free text comments written about university subjects are used as the dataset. We have compared two textual representations: a frequency word representation and term frequency relationship to word vectors, and found that word vectors provide a greater accuracy. However, these word vectors have a large number of features which aggravates the burden of computational complexity. Thus, we explored using a non-linear dependency measure for feature selection by maximizing the dependence between the text reviews and corresponding scores. Our quantitative and qualitative analysis on a student satisfaction dataset shows that our approach achieves comparable accuracy to the full feature vector, while being an order of magnitude faster in testing. These text analysis and feature reduction techniques can be used for other textual data applications such as sentiment analysis. 
\end{abstract}


% \fi
\IEEEpeerreviewmaketitle

\section{Introduction}
Statistical analysis techniques are used to find patterns in actual data. Statistical dependence measures such as Canonical Correlation Analysis (CCA) \cite{Bach2002}, Maximum Mean Discrepancy (MMD) \cite{Schneider2012}, and the Randomized dependence coefficient (RDC) \cite{lopez2013randomized} have been extensively used in the areas of pattern recognition and computer vision to find correlation between the random variables. Here, we make use of the state-of-the-art non-linear dependency measures for text classification. We first perform feature selection via maximising the dependence of the text comments and the correlated scores, and then train the classifier on the reduced features.

Text analysis methods and techniques are applied to many actual examples of data to obtain meaningful examples and results. Textual data can be used, when looking at product reviews to determine if that product has a positive or negative satisfaction score. Movie reviews consisting of solely text data have also been used to predict ratings or scores \cite{turney2002thumbs}. In this work, we use student comments, paired with satisfaction. In this study, machine learning is used to provide a link between the satisfaction score given to a university subject, and free text comment. 

Student evaluations of teaching are an important part of assessing student satisfaction in a particular class. The feedback from these student evaluations often consists of a score or ranking for several questions, and also the possibility of free text comments. Part of teaching is fulfilling students' expectations, and feedback and ratings from evaluation allows that to be done \cite{CheongCheng1997}. From feedback, pointers and information can be determined and hopefully influence the teaching style and method of delivery. 

In its most basic form, student satisfaction for a unit (subject) can be rated as either good, neutral or bad. Maas et al \cite{Maas2011} show a generic approach to using words as a vector and their application to sentiment analysis. The methods mentioned in this work look at algorithms and search to find patterns and common connections.

In this work we aim to investigate the following two research questions,
\begin{enumerate}
\item How can dimensions of data that provide the highest influence be determined?
\item What is the correlation between the text a student writes and the satisfaction score a student gives?
\end{enumerate}

Our contribution in this work is three-fold. We examine whether there is a dependence between students' free text comments and the satisfaction score which they give. We introduced a data-set, consisting of many scores and free text comments. We finally use statistical dependency methods, performing a quantitative and qualitative analysis on the results.

\section{Literature Review}
Text analysis is a large area with several different methods and techniques being commonly applied. The literature review discusses several common techniques, focussing on the ones used in this work. 
\subsection{Text Analysis and Bag of Words Representation}
Text data contains a plethora of ingrained or implied information and meaning, which humans are naturally adept at interpreting. This is however harder for machines to extract, often resulting in misinterpreted or entirely lost information which may be crucial to the context. One example of text data that is used often as a bench mark is movie reviews. A study, that provides valuable insights into text data analysis, looks at predicting movie revenue from ratings written by critics \cite{Joshi2010}.

Traditionally, a bag of words representation is used to model textual data. This involves transforming the corpus of text into an $N \times M$ sparse matrix, where $N$ is the number of text responses and $M$ is the number of unique words. 

Another common enhancement used alongside the bag of words representation is Term Frequency - Inverse Document Frequency (TF-IDF). TF-IDF calculates values for every word which is a proportion of the word frequency in one document with respect to the frequency percentage of all documents the word appears in \cite{Ramos2003}. The higher this number, suggests a greater importance for this word in the current document. Common words would receive a lower weighting. This can be also expressed in the following expression,

\begin{equation}
w_{x,y} = \text{tf}_{x,y} \times \log \left( \frac{N}{df_{x}} \right)
\end{equation}

Where, $w_{x,y}$ is the weight of term $x$ within document $y$, $\text{tf}_{x,y}$ is the frequency of $x$ in $y$, $df_{x}$ is the number of documents containing $x$, and $N$ is the total number of documents. 

\subsection{Vector Space Models}
An alternative way to represent textual data, is to use a vector of features for each individual word. These feature vectors have many more dimensions than just a single word, and when developed, are trained from many documents.
\subsubsection{Word2Vec}
Developed in 2013, Word2Vec is one method of modelling words as vectors \cite{Mikolov2013}. Other versions developed using similar processes exist, however this model was developed using a neural network training approach, and a negative skip gram model \cite{Goldberg2014}. Each word is projected into a $300 \times 1$ vector. This Word2Vec model has been trained on Google News articles.

\subsubsection{Glove}
Another alternative to Word2Vec is the word vector representation Glove \cite{Pennington2014}. This method uses a technique (global matrix factorization) to implement the word vectors differently to Word2Vec. This representation can sometimes outperform Word2Vec, depending on the situation. 
\subsubsection{t-SNE}
Visualising high-dimensional data is an important problem and needs to be considered carefully \cite{Heuer2015}. Using t-Distributed Stochastic Neighbour Embedding (t-SNE), data which previously has many dimensions can be reduced to just two or three for visualisation purposes \cite{VanDerMaaten2008}. It is useful when high dimensional data has common or important themes across only a few of the dimensions. t-SNE works well with larger amounts of data, as opposed to just a few vectors. 

\subsubsection{Summary}
Word2Vec and Glove are two examples of word vectorisation methods which are commonly used for text analysis. Representing words as vectors allows further embedded information to be uncovered, compared to more simplistic analysis techniques. 

\section{Background}
In this work, we are interested in measuring the distance between distributions of the student's comments and their corresponding scores. Generally, the two probability distributions can be compared either through non-parametric models (e.g., kernel density estimation), or parametric ones (e.g., using Gaussian Mixture Models).

Here, we exploit two non-parametric approaches to compute the distribution difference between multiple sources of data: MMD (Maximum Mean Discrepancy), and RDC (Randomised Dependence Coefficient).

\subsection{MMD (Maximum Mean Discrepancy)}
Let $\mat{X}_p = \{\vec{x}_p^1, \cdots, \vec{x}_p^n\}$ and $\mat{Y}_q = \{\vec{y}_q^1, \cdots, \vec{y}_q^n\}$ be the two sets of i.i.d. observations from  two different sources $p$ and $q$, with $n$ samples, respectively. Using the MMD criterion, we can determine whether $p = q$.

\begin{definition}
	\cite{Gretton2006} Let $\mathfrak{F}$ be a class of functions $f : \mathcal{X} \rightarrow \mathbb{R}$. Then the MMD and its empirical estimate are defined as:
	\begin{eqnarray}
	\operatorname{MMD}(\mathfrak{F},p,q)= \sup_{f \in \mathfrak{F}}\left(E_{x \sim p}[f(x)]-E_{y\sim q}[f(y)]\right),\nonumber\\
	\operatorname{MMD}(\mathfrak{F},\mat{X}_p,\mat{Y}_q)= \sup_{f \in \mathfrak{F}}\left( \frac{1}{n}\sum\limits_{i=1}^nf(\vec{x}_p^i)-\frac{1}{n}\sum\limits_{j=1}^nf(\vec{y}_q^j)\right). \nonumber
	\end{eqnarray}%
\end{definition}

\begin{exe}
	\cite{Gretton2006} Let $ \mathfrak{F}$ be a unit ball in a Reproducing Kernel Hilbert Space (RKHS), defined on a compact metric space $ \mathcal{X}$  with associated kernel $k(\cdot,\cdot)$. Then $\text{MMD}(\mathfrak{F},p,q)=0 $ if and only if $p=q$.
\end{exe}
%MMD is a powerful non-parametric method that computes the distribution distance between two sets of data by mapping the data to reproducing Kernel Hilbert Space (RKHS).

An empirical estimate of the MMD can be written as $\left(\sum\limits_{i,j=1}^n{\dfrac{k(\Vec{x}_p^i,\Vec{x}_p^j)}{n^2}} + \sum\limits_{i,j=1}^n {\dfrac{k(\Vec{y}_q^i,\Vec{y}_q^j)}{n^2}} -2 \sum\limits_{i,j=1}^{n,n}{\dfrac{k(\Vec{x}_p^i,\Vec{y}_q^j)}{n^2}}\right)^{\frac{1}{2}}$ where $k(\cdot,\cdot)$ is the universal or more general form of the characteristic kernel of the mapping:
\begin{eqnarray}
&&\hspace{-1.4cm}\operatorname{MMD} = \| \frac{1}{n}\sum\limits_{i=1}^n\phi(\vec{x}_p^i)-\frac{1}{n}\sum\limits_{j=1}^m \phi (\vec{x}_q^j) \|^2 \nonumber\\
&&\hspace{-0.4cm} = \frac {1}{n^2} \sum_{i,j=1}^n\exp\left(-\frac{(\vec{x}_p^i-\vec{x}_p^j)^T  (\vec{x}_p^i-\vec{x}_p^j)}{\sigma}\right)\nonumber\\
&&\hspace{-0.4cm} +\frac {1}{n^2}\sum_{i,j=1}^n \exp\left(-\frac{(\vec{y}_q^i-\vec{y}_q^j)^T  (\vec{y}_q^i-\vec{y}_q^j)}{\sigma}\right) \label{eqn:mmd_gauss}\\
&&\hspace{-0.4cm}  -\frac {2}{n^2}\sum_{i,j=1}^{n,n}  \exp\left(-\frac{(\vec{x}_p^i-\vec{y}_q^j)^T (\vec{x}_p^i-\vec{y}_q^j)}{\sigma}\right). \nonumber
\label{eqn:eq1}
\end{eqnarray}
In summary,  MMD is a powerful non-parametric method that compares the distribution difference between different sources of data by computing the difference between the means of the two sets mapped into a RKHS.

\subsection{RDC (Randomized Dependence Coefficient)}
The \emph{Randomized Dependence Coefficient} (RDC) statistic measures the dependence between random samples $\bm X \in \mathbb{R}^{p\times n}$ and $\bm Y \in \mathbb{R}^{q\times n}$ by first applying a copula-transformation on the random samples and projecting  the copulas through $k$ randomly chosen non-linear projections, and then finding the largest canonical correlation between these non-linear projections. Given the random samples $\bm X \in \mathbb{R}^{p\times n}$ and $\bm Y \in
\mathbb{R}^{q\times n}$ and the parameters $k \in \mathbb{N}_+$ and $s \in
\mathbb{R}_+$, the Randomized Dependence Coefficient between $\bm X$ and $\bm
Y$ is defined as: \small
\begin{equation}\label{eq:rdc}
\text{rdc}(\bm X, \bm Y; k,s) :=
\sup_{\bm \alpha, \bm \beta}\rho\left(
\bm \alpha^T \bm \Phi(\bm P(\bm X); k, s),
\bm \beta^T \bm \Phi(\bm P(\bm Y); k, s)\right).
\end{equation}
\normalsize
where  $\bm \Phi(\bm X; k,s)$ is the $k-$th order random non-linear projection from $\bm X \in \mathbb{R}^{d
	\times n}$ to $\bm \Phi(\bm X; k, s) \in
\mathbb{R}^{k \times n}$:
\begin{equation}
\bm \Phi(\bm X; k,s) := 
\left(
\begin{array}{ccc}
\phi(\bm w_1^T \bm x_1+b_1) & \cdots & \phi(\bm w_k^T \bm x_1+b_k)\\
\vdots & \vdots & \vdots \\
\phi(\bm w_1^T \bm x_n+b_1) & \cdots & \phi(\bm w_k^T \bm x_n +b_k)
\end{array}
\right)^T
\end{equation}
and $\rho$ is the largest canonical correlation between the non-linear projections $\bm \alpha^T
\bm X$ and $\bm \beta^T \bm Y$ of two random samples $\bm X \in
\mathbb{R}^{p\times n}$ and $\bm Y \in \mathbb{R}^{q\times n}$.

We used the Randomized Dependence Coefficient (RDC) as a measure of dependence between the student's comments and their corresponding scores. RDC is defined as the largest canonical correlation between random non-linear projections of the variables' copula transformations. Unlike the other non-linear dependence measures such as Kernel Canonical Correlation Analysis (KCCA) \cite{Bach2002} and Copula Maximum Mean Discrepancy \cite{Schneider2012} which exhibit prohibitive running times on large scale data, RDC has low computational cost of $O(n \log n)$, where $n$ is the number of samples. Moreover, RDC is easy to implement, invariant to monotonically increasing transformations, and performs well under the existence of additive noise.

\section{Proposed Approach}

In this work a combination of word vectors and feature reduction techniques are used. This section will discuss how these methods were applied. 
\subsection{Word2Vec}

Word2Vec is one method for training word vectors. The Google News data set contains word vectors which were trained on around 100 billion words from Google News articles. Some other common models are trained on data sets such as Wikipedia, but the Google News set is more widely used. 

Figure \ref{w2v_model} below shows the neural network model in which Word2Vec is trained. The network has a single hidden layer. Both the output and output layers consist of the number of neurons equal to the number of unique words in the training vocabulary. The number of neurons in the hidden layer is equal to the dimentionality of the word vector \cite{Sugathadasa2017}.

\begin{figure}[h]
\centering
\includegraphics[width=6cm]{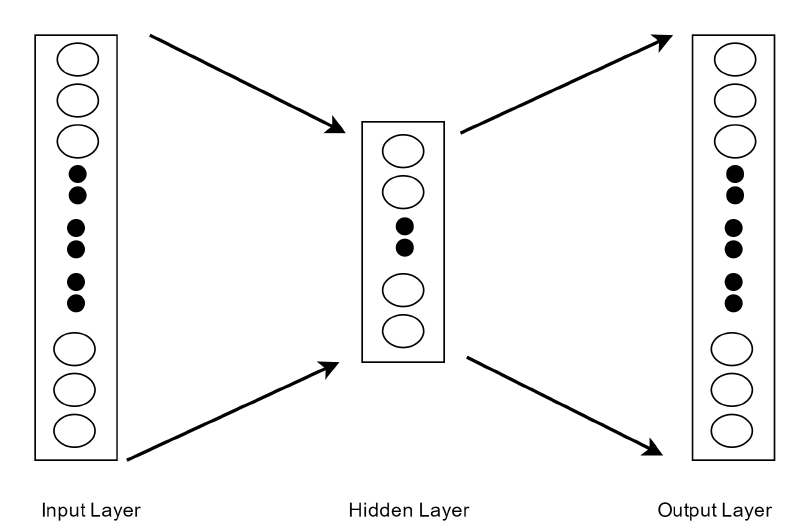}
% where an .eps filename suffix will be assumed under latex, 
% and a .pdf suffix will be assumed for pdflatex; or what has been declared
% via \DeclareGraphicsExtensions.
\caption{Word2Vec Model \cite{Sugathadasa2017}}
\label{w2v_model}
\end{figure}

\begin{figure}[htp]
\centering
\includegraphics[width=0.45\textwidth]{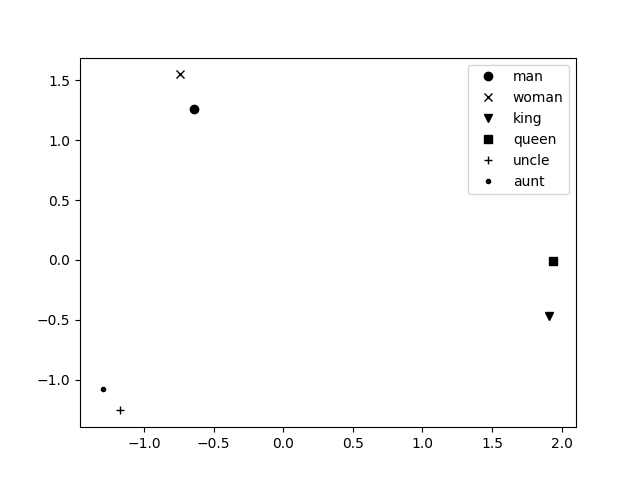}
% where an .eps filename suffix will be assumed under latex, 
% and a .pdf suffix will be assumed for pdflatex; or what has been declared
% via \DeclareGraphicsExtensions.
\caption{Example of Reduced Word Vector}
\label{w2v_sim}
\end{figure}

Figure \ref{w2v_sim} shows an example of how Word2Vec can model particular words once reduced to 2 dimensional space (in this case using PCA). You can see relationships between certain words represented in a vector form. As an equation, we can express the relationship between these vectors in the following equation,

\begin{equation}
V(\text{king}) - V(\text{man}) + V(\text{woman}) \approx V(\text{queen})
\end{equation}

This equation describes a gender relationship, which has been learnt through embedded neural networks. As a further example, finding the angle between two vectors demonstrates the similarity between those two words. 

Another example of a gender relationship demonstrated in Figure \ref{w2v_sim} is the equation below. 

\begin{equation}
V(\text{uncle}) - V(\text{man}) + V(\text{woman}) \approx V(\text{aunt})
\end{equation}
 
These are just two examples of equations which demonstrate the implicit understanding of the Word2Vec model trained on many articles. 
 
\subsection{Choosing Dimensionality}
We validated the empirical performance of RDC and MMD on the student's evaluation data of university subjects.  The random dimension for RDC was set to $20$ as we could not observe any improvements for embedding dimensions $d \geq 100$. Moreover, as noted in \cite{Oymak2015}, the randomised embedding dimension should not be too much smaller than the original dimension $D$ to prevent a point in the set from being mapped to the origin. To ensure this, we refer to the Universality Law for the Embedding Dimension from \cite{Oymak2015,Erfani2016}: 

\begin{exe}
	{\bf A Universality Law for the Embedding Dimension}.
	Given the $n \times h$ random projector $\phi$ with the parameters $p > 4$, $\nu \geq 1$, $\varrho \in (0,1)$, and $\varepsilon \in (0,1)$, there is a number $N := N(p,\nu,\varrho,\varepsilon)$ for which the following statement holds. Suppose that the ambient dimension $n \geq N$; $E$ is a nonempty, compact subset of $\mathbb{R}^n$ that does not contain the origin; the statistical dimension of $E$ is proportional to the ambient dimension: $\varrho n \leq \theta(E) \leq n$. Then $h \geq (1+\varepsilon) \theta(E)$ implies $\mathbb{P} \{ 0 \notin \phi(E) \} \geq 1 - C_p n^{1 - \frac{p}{4}}$. Furthermore, if $\theta(E)$ is spherically convex, then
	$h \leq (1 - \varepsilon)\theta(E)$ implies $\mathbb{P} \{ 0 \in \phi(E) \} \geq 1 - C_p n^{1 - \frac{p}{4}}$.
\end{exe}

We performed greedy word selection to construct the subset of words in the students comments that maximises the dependence between the word set and the associated score value. We reported the results of our algorithm as a set of words which has the maximum dependency to the target value.

The computational complexity of RDC is $O((p+q)n \log n+kn \log(pq)+k 2n)$ where $p$ and $q$ are the dimensions of the random variables, $n$ is the sample size, and $k$ is the reduced dimension. The cost of RDC can be approximated by $O(n \log n)$ if applied on large scale data (very large $n$).

\subsection{Feature Reduction}

The proposed approach for feature reduction uses a greedy search method for calculating correlation with both RDC and MMD. This has also been compared to PCA for reference. It was decided that the features for each vector would be reduced from 300 to 20 using each method. For both RDC and MMD, we started with 1 feature for all words ($ N \times 1$), and found the maximum correlation between a feature, and the output label. After the optimum single feature was chosen, this was then repeated, adding another feature to the input, and maximising that correlation. This was repeated, until the 20 features with the combined highest correlation where selected. Figure \ref{corr} below shows an example of this graphically. In this figure, $N$ is the number of text responses, $C$ is the correlation, and $i$ is the number of features being tested (between 1 and 20). The function being used is either the RDC or MMD feature correlation. 

\begin{figure}[H]
\centering
\includegraphics[width=5cm]{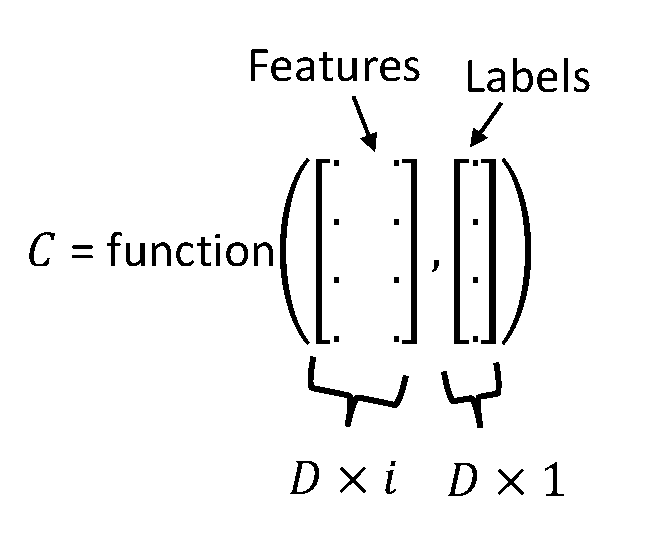}
\caption{Correlation Calculation}
\label{corr}
\end{figure}

This was designed to maximise the correlation, adding one feature at a time. 

\section{Experimental Results}
\subsection{The Data Set}
The data for this experiment was gathered from the Queensland University of Technology. QUT runs two student surveys titled `Pulse' and `Insight' as a part of the framework `Reframe' for evaluating learning and teaching \cite{qut2013}. The first survey, the Pulse survey, solicits students' feedback in the early weeks of the semester. The second survey, titled Insight, surveys students at the end of semester. In each of these surveys, students are asked to rate their views on three statements, however we focus on the final question ``I am satisfied with this unit so far."

Each statement was responded to on the Likert scale, 1 being disagree strongly, and 5 being
agree strongly. After answering these questions, an open ended response was left optional,
for students to respond to with feedback and suggestions asking them to ``Please provide any
further feedback you may have about this unit.". This feedback provides recommendations
and suggestions for teaching staff to read and take into consideration.

This study and analysis used Pulse and Insight survey data from 19 units
in the Science and Engineering Faculty, over a 4-year duration. The data consisted of more than 3000 responses which included the numerical and text feedback.

Examples of some responses include,

\begin{itemize}
\item Satisfaction rating: 4. Free response: \textit{``Excellent work. Very helpful staff and lots of assistance given via video tutorials."}
\item Satisfaction rating: 5. \textit{``Great structured unit. Very well organised and great learning environments"}
\item Satisfaction rating: 1. Free response: \textit{``The unit moved too quickly through concepts, and the content in the workshops has not helped with assessment items"}
\item Satisfaction rating: 2. Free response: \textit{``The assessment and the lectures didn't relate well - very confusing overall. Not enough practice problems."}
\end{itemize}

\subsection{Procedure}
The procedure was broken down into four major steps: Data Preparation and Preprocessing, Removing excess data, Reducing dimensionality and applying various machine models.
\subsubsection{Data Preparation and Preprocessing}
The text data for each student response was initially preprocessed. This involves:
\begin{itemize}
\item Turning all text into lower case.
\item Remove punctuation from the text data.
\item Remove stop words (i.e. `the', `and', `it')
\item Split data into individual words, instead of one sentence.
\end{itemize}
After this, the pre-existing Word2Vec model, trained from news articles was applied to each individual student response. Each word in the response was transformed into a vector, and the average of all these vectors taken. This vector was then normalised before further analysis was undertaken.

Initial results, as well as manual inspection determined that being able to tell the difference between strongly agree (5/5) and agree (4/5), even for the human eye can be difficult. The classes were then rounded from the original 5 categories into 3 smaller categories (Disagree, Neutral and Agree). 

\begin{enumerate}
\item Category 1 - 1 or 2 out of 5 (Disagree)
\item Category 2 - 3 out of 5 (Neutral)
\item Category 3 - 4 or 5 out of 5 (Agree)
\end{enumerate}

\subsubsection{Remove Excess Data}
Initial work showed higher accuracies, however it was noticed that much of the data was unbalanced - many more agree responses. This was unfortunately giving predictions which didn't match expectations. The data was therefore levelled out to have similar numbers in all three chosen classes

\subsubsection{Reduce Dimensionality}
In several of the experiments, we used dimension reducing techniques (PCA, MMD and RDC) to reduce the complexity of the data. Following the procedure discussed previously in Section IV, the 300 dimension feature vectors were reduced down to be 20 dimensions. 

\subsubsection{Apply Models}
Several common machine learning models were selected to evaluate the accuracy of the methods being testes. These methods are listed below. Several of these methods also required additional parameters which were configured. 
\begin{itemize}
\item Extra Trees Classifier (Etrees)
\item Linear Discriminant Analysis (LDA) - Using an SVD solver
\item Logistic Regression (Log) - Regularisation strength value of $C=1$
\item k-Nearest Neighbour (kNN) - Using the 5 nearest neighbours, $k=5$, and a uniform weighting
\item Decision Tree Classifier (DT)
\item Gaussian Naive Bayes (G-NB)
\item Support Vector Machine - Linear (L-SVM)
\item Support Vector Machine - Gaussian (G-SVM) - RBF kernel used, with a Regularisation strength value of $C=1$ and the variance $\sigma$ of the Gaussian kernel is systematically chosen to be the median squared distance between all training samples.
\item Support Vector Regression - Linear (L-SVR) - since the output of this method is continuous, results were rounded to the nearest class. 
\item Support Vector Regression - Gaussian (G-SVR) - since the output of this method is continuous, results were rounded to the nearest class. RBF kernel used, with a Regularisation strength value of $C=1$ and the variance $\sigma$ of the Gaussian kernel is systematically chosen to be the median squared distance between all training samples.
\end{itemize}

After configuring each of these models, they were applied in a 5-fold validation, with an 80/20 training test data split.

\subsection{Quantitative Results}
\subsubsection{Bag of Words and TF-IDF Accuracies}
Bag of words and TF-IDF are used first to get an initial accuracy. Several common machine learning algorithms were picked as initial benchmarks for testing. 

\begin{table}[H]
\renewcommand{\arraystretch}{1.3}

\centering
\caption{Bag of Words and TF-IDF Results}
\label{bow_res}
\begin{tabular}{|l|l|l|}
\hline
\begin{tabular}[c]{@{}l@{}}Machine Learning \\ Method\end{tabular} & \begin{tabular}[c]{@{}l@{}}BOW \\ Accuracy (\%)\end{tabular} & \begin{tabular}[c]{@{}l@{}}TF-IDF \\ Accuracy (\%)\end{tabular} \\ \hline
Etrees & 46.08                                                        & 45.07                                                           \\ \hline
LDA                                       & 39.08                                                        & 40.54                                                           \\ \hline
Log                                                & 47.83                                                        & 43.77                                                           \\ \hline
k-NN                                                 & 35.86                                                        & 33.08                                                           \\ \hline
DT                                                      & 43.31                                                        & 43.48                                                           \\ \hline
G-NB                                               & 46.53                                                        & 45.46                                                           \\ \hline
\textbf{L-SVM} & 46.96                                                        & \textbf{47.21                                                          } \\ \hline
G-SVM                                  & 36.76                                                        & 42.35                                                           \\ \hline
L-SVR & 37.89                                                        & 43.53                                                           \\ \hline
G-SVR & 35.91                                                        & 40.93                                                           \\ \hline
\end{tabular}
\end{table}

These results here show initial promising accuracies, especially for the SVMs - the TF-IDF slightly outperforming the BOW method. 

\subsubsection{Word2Vec Accuracies}
Representing the words as vectors, allows more information to be uncovered. These results show Word2Vec with the same machine learning models that were tested above. 

Since the Word2Vec model has a high dimensionality space (300), different methods were explored to reduce the complexity of the data. These vectors were reduced to 20 dimensions, following the procedure mentioned previously. Table \ref{w2v_res} shows some accuracies from both sets of methods. 

\begin{table}[htb]
\renewcommand{\arraystretch}{1.3}
\centering
\caption{Word2Vec Model Results}
\label{w2v_res}
\begin{tabular}{|l|l|}
\hline
Machine Learning Method              & Accuracy (\%) \\ \hline 
W2V + Etrees                & 45.52         \\ \hline
W2V + LDA         & 49.97         \\ \hline
W2V + Log                  & 49.18         \\ \hline
W2V + k-NN                  & 42.46         \\ \hline
W2V + DT                        & 40.37         \\ \hline
W2V + G-NB                 & 48.11         \\ \hline
W2V + L-SVM      & 50.65         \\ \hline
\textbf{W2V + G-SVM}    & \textbf{51.67}         \\ \hline
W2V + L-SVR   & 44.27         \\ \hline
W2V + G-SVR & 45.90         \\ \hline \hline
W2V + PCA + G-SVM & 47.77         \\ \hline
\textbf{W2V + RCD + G-SVM} & \textbf{45.51}         \\ \hline
W2V + MMD + G-SVM & 36.09         \\ \hline
\end{tabular}
\end{table}

The initial results from Table \ref{w2v_res} show that the Gaussian Kernel - SVM produced the highest accuracy (51.67\%) in predicting student satisfaction scores. The accuracy was also higher than the best accuracy from the bag of words and TF-IDF methods demonstrated above. 

We can also see that in terms of accuracy the reduced dimension vectors (20 dimensions) do not perform as well as the full 300 dimension vectors. They however, do have significantly less computational cost. PCA performs best from the compared techniques, and the accuracy of MMD is significantly less. However, comparing the computational cost of RDC against PCA, RDC is smaller. This demonstrates the trade-off between accuracy and complexity. RDC and MMD were used as state-of-the-art methods, and PCA used as a comparison point. 

\subsection{Qualitative Results}
Previous results sections above have looked at the accuracies of individual methods, in this section we focus on several selected comments in further detail. In this section Pos refers to positive, Neu to negative and Neg to negative. 

\subsubsection{Correctly Classified Responses}

The examples in this first section show examples of comments that were classified correctly by the Gaussian SVM model, as well as all three tested feature reduction techniques. 

\begin{table}[H]
\centering
\caption{Example 1 - Negative Review - All Correct}
\label{ex1}
\begin{tabular}{|c|c|c|c|}
\hline
Gaussian SVM & RDC          & PCA          & MMD          \\ \hline
\cmark & \cmark & \cmark & \cmark \\ \hline
Neg & Neg & Neg & Neg \\ \hline
\end{tabular}
\end{table}

Response: \textit{``Not having the answers available from tutorials I felt was a disadvantage. This is an introductory course, it's not like we are 4th year electrical students. When struggling to learn the content and work out questions and not having a concrete answer to check was very demotivating. The people who would abuse having answers available and not do the work for them selves are the ones that wouldn't do the work in the first place. I felt like the tutorial class sizes were too big and it would of been on benefit to have smaller groups to encourage discussion. Assessment so far has been set out well."} 

This first example above was rated as a ``negative" result by a student, which is also clear to the reader when examining it as well. The Gaussian SVM, as well as the three different feature reduction methods (RDC, MMD and PCA) all predict this to be a positive result also. This demonstrates the ability for Word2Vec to operate on long responses, even though the average is computed for all the words in the response. The feature reduction methods also retain the important dimensional vectors for classifying the response.

\begin{table}[H]
\centering
\caption{Example 2 - Positive Review - Most Correct}
\label{ex2}
\begin{tabular}{|c|c|c|c|}
\hline
Gaussian SVM & RDC          & PCA          & MMD          \\ \hline
\cmark & \cmark & \cmark & \xmark \\ \hline
Pos & Pos & Pos & Neg \\ \hline
\end{tabular}
\end{table}

Response: \textit{``This unit is presented well by (Lecturer Name).  He is enthusiastic and his teacher/student interaction is good.  The only very minor issue that I may have is that his examples are sometimes hard to read with his handwriting, especially if he is using a thick point pen, like a white board marker.  All in all the unit is good and he explains it all well."} 

This example was rated ``positive", once again showing correlation between the satisfaction score given, as well as the scores predicted by the algorithms. The only discrepancy here was the prediction from MMD. This however is expected, as the overall accuracy for MMD was much lower than the PCA or RDC methods.

\begin{table}[H]
\centering
\caption{Example 3 - Positive Review - All Correct}
\label{ex3}
\begin{tabular}{|c|c|c|c|}
\hline
Gaussian SVM & RDC          & PCA          & MMD          \\ \hline
\cmark & \cmark & \cmark & \cmark \\ \hline
Pos & Pos & Pos & Pos \\ \hline
\end{tabular}
\end{table}

Response: \textit{"Good unit."} 

This result, similar to the previous one was rated as ``positive" by the student. This was correctly classified by the Gaussian SVM, as well as all three feature reduction methods. This example was included to demonstrate that even short responses of only a few words can still be correctly classified.

\subsubsection{Incorrectly Labelled Responses}

Included in this section are several responses that were 'classified incorrect', however upon further inspection, are perhaps underlying difficulties associated with the data set. 

\begin{table}[H]
\centering
\caption{Example 4 - Neural Review That Should be Negative}
\label{ex4}
\begin{tabular}{|c|c|c|c|}
\hline
Gaussian SVM & RDC          & PCA          & MMD          \\ \hline
\cmark & \xmark & \cmark & \cmark \\ \hline
Neu & Neg & Neu & Neu \\ \hline
\end{tabular}
\end{table}

Response: \textit{``Lectures are not engaging. Feels like either way too much content to go through for 2 hours lecture pace is a bit fast  tutorials are good"} 

Reading this review here, it appears that the students comments are mostly negative, however this was marked as ``Neutral" by the corresponding student. The G-SVM, PCA and MMD methods all classified this response as ``Neutral", but it was marked negative by the RDC reduction method. Although this response was technically classified incorrect by RDC, the we believe that the RDC chosen category is more suitable.  

\begin{table}[H]
\centering
\caption{Example 5 - Positive Review That Should be Neutral}
\label{ex5}
\begin{tabular}{|c|c|c|c|}
\hline
Gaussian SVM & RDC          & PCA          & MMD          \\ \hline
\xmark & \xmark & \xmark & \xmark \\ \hline
Neg & Neu & Neg & Neu \\ \hline
\end{tabular}
\end{table}

Response: \textit{``Lecturer has to explain things a bit better as most students are currently completing (Unit Name) at the same time."} 

Again, this is another example which the RDC reduced feature vector produces a better representation. The rating given by the student was ``positive", however reading the comment, it appears to be most likely Neutral, or slightly negative. Both RDC and MMD scored this comment as ``Neutral", which we believe is more indicative of the comment.

\subsection{Summary}
The selected comments above are a demonstration of some examples where the chosen algorithms have correctly identified the label category, and examples of where this is not the case. This demonstrates RDC being able to select relevant feature vectors from the many provided by Word2Vec. 

The qualitative results also demonstrate the importance of not solely relying on accuracy scores for these comments, especially since the satisfaction score and students comments may not line up, or give all relevant information. 

Both the qualitative and quantitative results highlight different points of importance. The quantitative results demonstrate the benefit of modelling the students responses as vectors, and show that the Gaussian SVM returns the best classified results. We also see however, that examining some of the results further in detail, feature reduction techniques, mainly demonstrate classification which perhaps follows more closely to

\section{Conclusions and Future Work}
In this work, we have compared several text analysis techniques in the context of predicting student satisfaction. We have demonstrated a strong correlation between student satisfaction scores for a particular unit. Word2Vec provided the best accuracy in predicting satisfaction when using a Gaussian SVM model. The Word2Vec model used contained a large number of dimensions, so we also explored several feature reduction techniques. PCA, as expected, provided the best numerical accuracy of the three tested techniques (PCA, RDC and MMD), however RDC was a close second. Looking at several responses qualitatively, RDC was shown to obtain results which were not necessarily classified correctly, but when reading the response manually, seemed to fit more into the category which RDC placed it in. These techniques allow those dimensions which have the highest influence to be found.

This work provides a comparison between various machine learning models and feature reduction techniques. A strong correlation was shown between the text written by a student, and the score that they give. The initial accuracy provides a promising result, but there are several ways that we believe the performance could be improved. 

Dictionary learning could be an alternative approach to improving accuracy. Learning a specific dictionary for this task would allow important words to be selected, and means that ones which are less relevant could be filtered out. Words that do not impact the score, need not be considered. 

Finally, one aspect that is lacking from the model currently being used is word order. When the word vector is calculated the vectors are averaged across the sentence. This means that a sentence with the same words in a different order would have the same word vector. Using a Recurrent Neural Network as part of the process would allow word order within sentences to become a part of the process as well. 

This work shows the application of these techniques on one set of textual data, and the corresponding scores, however the procedures shown can be applied on various other datasets. 


\section*{Acknowledgement}

The authors would like to thank QUT for providing the data, and QUT students for completing the survey.



\bibliographystyle{IEEEtran}
\bibliography{references}

\end{document}